%% file: main.tex
\documentclass[runningheads]{llncs}

\usepackage{wrapfig}
\usepackage{multirow}
\usepackage[normalem]{ulem}
\useunder{\uline}{\ul}{}
\usepackage[T1]{fontenc}

\usepackage{algorithm}
\usepackage{algorithmic}

\usepackage{eccv}

\usepackage{eccvabbrv}

\usepackage{graphicx}
\usepackage{booktabs}

\usepackage[accsupp]{axessibility}  

\usepackage{hyperref}

\usepackage{orcidlink}

\begin{document}

\title{FoundDP: Revisiting Weak Disparity Observability in Dual-Pixel Depth Estimation} 





\titlerunning{FoundDP}

\author{
Fengchen He\orcidlink{0009-0000-8152-5703}\textsuperscript{*} \and
Hao Xu\orcidlink{0009-0006-6803-5463}\textsuperscript{*} \and
Dayang Zhao \and
Tingwei Quan \and
Shaoqun Zeng
}

\institute{
    Huazhong University of Science and Technology, Wuhan, China \\
    \textsuperscript{*}Equal contribution \\
    \email{{linyark, xuhao, dayangzhao, quantingwei, sqzeng}@hust.edu.cn} 
}

\authorrunning{F.~He et al.}

\maketitle
\input{main_sec/0_abstract}    
\input{main_sec/1_intro}

\input{main_sec/2_re}
\input{main_sec/3_method}
\input{main_sec/4_exp}
\input{main_sec/5_conc}

\section*{Acknowledgments}
This work was supported by National Natural Science Foundation of China (Grant No. 32471146) and the project N20240194.
The authors thank Echossom, Miya, and Xinge for valuable discussions and assistance. 

\bibliographystyle{splncs04}
\bibliography{refs}

\end{document}

%% file: main_sec/0_abstract.tex
\begin{abstract}
Dual-pixel (DP) imaging enables metric depth estimation from a single camera using sub-aperture disparity.
However, the extremely small effective baseline limits disparity observability, leading to structural degradation and depth failure in textureless, low-contrast, or downsampled regions.
Existing DP-based methods rely primarily on local disparity cues and therefore become unreliable when disparity signals are weak or ambiguous.
To address this limitation, we propose \emph{FoundDP}, a unified framework that integrates metric DP depth with global structural priors from a monocular depth foundation model.
Our method preserves metric scale through DP-derived depth and leverages Vision Transformer (ViT) features to restore structural consistency in weak-disparity regions.
To ensure reliable metric guidance under DP imaging conditions, we identify and mitigate ViT representation degradation induced by DP defocus blur via ViT feature alignment, enabling stable metric-guided depth estimation.
Extensive experiments on synthetic and real-world DP benchmarks show that FoundDP delivers superior performance, with consistent gains in structural fidelity and metric accuracy, especially under reduced disparity observability.
Code will be available at: \url{https://github.com/EchoLighting/FoundDP}
\keywords{Dual-Pixel \and Metric Depth \and Depth Guidance}
\end{abstract}

%% file: main_sec/1_intro.tex
\section{Introduction}
\label{sec:intro}

Dual-pixel (DP) imaging~\cite{kobayashi2016low,shi2024split} equips a single camera with implicit stereo capability by splitting each pixel into left and right sub-pixels that capture light from slightly different viewpoints. Originally introduced for phase-detection autofocus~\cite{choi2023exploring,sliwinski2013simple}, this hardware design enables disparity acquisition without requiring additional sensors or explicit stereo baselines~\cite{kobayashi2016low}. 
Due to its minimal manufacturing overhead and widespread deployment in consumer devices, DP technology has become a practical hardware primitive for computational imaging, supporting applications such as image synthesis~\cite{pan2021dual,li2023learning,he2025simulating}, image deblurring~\cite{li2025learning,yang2023k3dn,abuolaim2022improving,yang2024ldp}, reflection removal~\cite{punnappurath2019reflection,yu2025enhanced}, and rain-drop removal~\cite{li2024dual}. 
More recently, learning-based approaches have leveraged the implicit sub-aperture disparity in DP images for metric depth estimation~\cite{garg2019learning,pan2021dual,xin2021defocus,he2025simulating} and disparity recovery~\cite{kim2023spatio,yu2025all,monin2024continuous}, enabling physically grounded depth inference from a single exposure.

\begin{figure*}[ht]
    \centering
    \includegraphics[width=\linewidth]{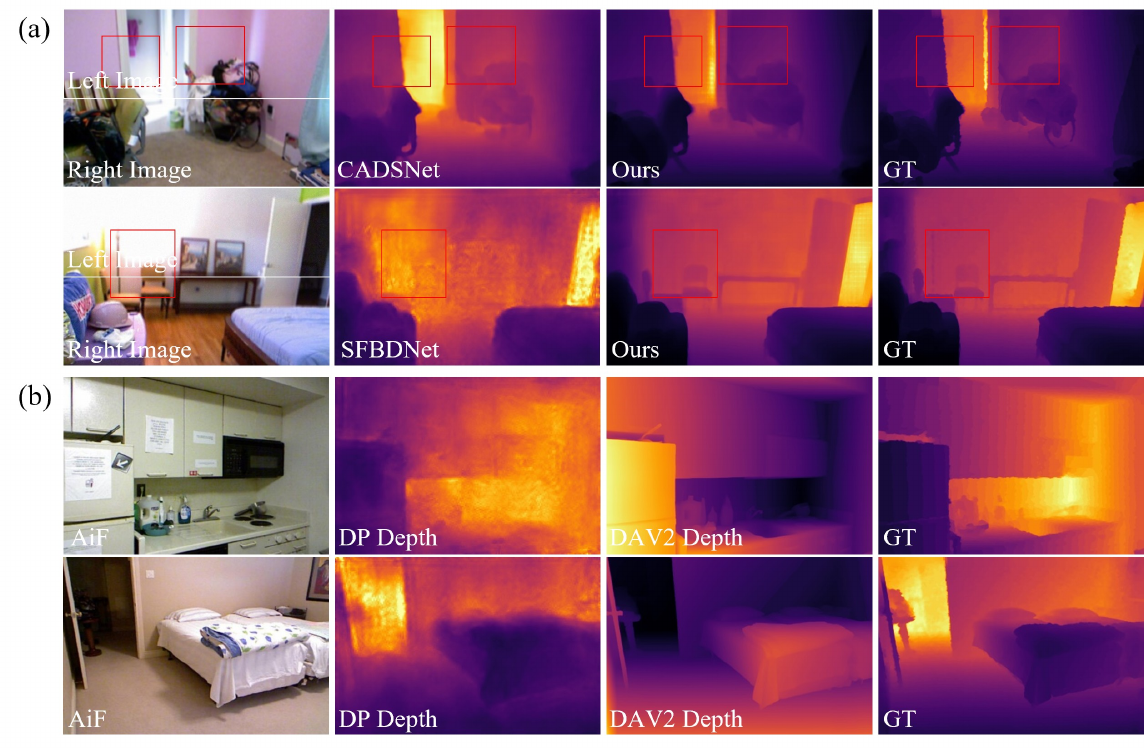}
    \vspace{-6mm}
    \caption{\label{fig:i_3}(a) Qualitative comparison. Our method yields sharper boundaries and more accurate depth in weak-disparity regions (highlighted in red).
    (b) Complementary failure modes: DPNet preserves metric scale but degrades structurally, while DAV2 recovers structure but lacks metric scale.
    Best viewed by zooming in.
    }
    \vspace{-2mm}
\end{figure*}

Although DP imaging can be modeled as a stereo system with a narrow effective baseline, its extremely small baseline fundamentally limits disparity observability.
Existing DP-based methods predominantly regress depth from local sub-pixel correspondence~\cite{pan2021dual,xin2021defocus}, which performs reliably in texture-rich and edge-distinct regions where disparity cues are observable.
However, in textureless, low-light, planar, or distant scenes, correspondence signals weaken toward the noise floor~\cite{wadhwa2018synthetic,he2025simulating}, leading to unreliable disparity estimation and unstable depth predictions, as illustrated in Fig.~\ref{fig:i_3}(a).
Furthermore, spatial downsampling reduces sub-pixel disparity precision, further weakening disparity observability and making depth estimation more unstable.
Since metric DP depth estimation depends on observable disparity~\cite{garg2019learning}, reduced observability inevitably causes structural degradation and depth failure, constituting a fundamental bottleneck for existing DP-based methods.

In contrast, recent monocular depth foundation models based on Vision Transformers (ViT)~\cite{yang2024depth,ranftl2021vision,wang2025moge,piccinelli2024unidepth} demonstrate remarkable structural reasoning and cross-scene generalization through large-scale pretraining. 
Models such as Depth Anything V2 (DAV2)~\cite{yang2024depthV2} and MoGe~\cite{wang2025moge} recover globally coherent scene geometry even in textureless or low-contrast regions by modeling long-range contextual dependencies. 
However, monocular predictions are statistically inferred rather than physically constrained, and remain ambiguous up to an affine transformation due to the absence of metric observability~\cite{ranftl2021vision}. 
As shown in Fig.~\ref{fig:i_3}(b), although foundation models maintain structural consistency, they lack reliable metric scale.

These observations expose a fundamental tension between physical observability and structural reasoning. 
DP imaging provides metric grounding but relies heavily on locally observable disparity, whereas foundation models offer strong global structural priors without physical scale constraints. 
This raises a key question: \emph{can metric depth estimation benefit from foundation-scale structural priors without sacrificing physical consistency?}

Motivated by this, we propose \emph{FoundDP}, a framework that leverages disparity observability to reconcile physically grounded DP cues with foundation-based structural priors for metric depth estimation. 
Our approach anchors metric scale through DP-derived depth while leveraging ViT-extracted global representations to restore structural continuity in weak-disparity regions. 
Specifically, we first obtain an initial metric depth from a DP network, adopt a ViT encoder to extract globally consistent structural features, and then apply depth guidance to obtain the final predictions.
This design enhances structural fidelity under weak-disparity conditions while preserving metric consistency.

However, directly applying foundation models to DP images is non-trivial. 
Due to optical defocus, DP images exhibit significant blur in out-of-focus regions~\cite{abuolaim2020defocus}. 
Compared with the sharp natural images used for foundation model pretraining, 
defocus introduces frequency attenuation and distribution shift in DP images, 
which systematically affects ViT representations and leads to inconsistent global attention responses during depth guidance.
We identify this as a previously underexplored source of guidance instability and introduce a ViT feature alignment strategy.
By enforcing feature consistency between paired clear images and their defocus-degraded counterparts, 
we align ViT features toward the feature distribution of ideal sharp inputs, thereby improving the stability of depth guidance.

We conduct extensive evaluations on multiple public synthetic~\cite{silberman2012indoor} and real-world DP datasets~\cite{punnappurath2020modeling,garg2019learning,li2023learning}, and additionally capture a dataset to analyze robustness under disparity attenuation. 
Results show that FoundDP consistently outperforms prior DP methods~\cite{garg2019learning,pan2021dual,kim2023spatio,ghanekar2024passive} in both normal and weak-disparity regions, validating the effectiveness of metric-guided depth estimation with foundation-based structural priors.

The main contributions of this work are summarized as follows:
\begin{itemize}

    \item We propose \emph{FoundDP}, a disparity-observability-aware framework that integrates metric DP depth cues with monocular foundation models to improve depth estimation under weak-disparity conditions.

    \item We identify ViT feature degradation under DP defocus blur as a key barrier to depth guidance and employ a feature alignment strategy to restore consistency for reliable metric-guided depth estimation.
    
    \item Extensive experiments on multiple public DP benchmarks demonstrate that FoundDP achieves superior performance, with consistent and often large gains in weak-disparity and reduced-observability conditions.
    
\end{itemize}

%% file: main_sec/2_re.tex
\section{Related Work}
\label{sec:re}

\subsection{DP-based Depth Estimation}

Since the introduction of DP sensors in consumer cameras, 
their potential for single-exposure depth recovery has been widely explored. 
DP imaging records signals from left and right sub-apertures at each pixel location, 
which can be interpreted as an extremely small-baseline stereo system~\cite{pan2021dual}. 
Wadhwa et al.~\cite{wadhwa2018synthetic} explicitly separated DP images into sub-views 
and applied conventional stereo matching, validating the implicit stereo nature of DP imaging 
and its feasibility for depth estimation. 
These physically grounded approaches reveal the intrinsic coupling among DP disparity, focus depth, and defocus blur, 
but remain highly sensitive to imaging conditions.

To improve robustness, learning-based methods were introduced. 
Garg et al.~\cite{garg2019learning} proposed an inverse depth estimation framework with affine-invariant constraints. 
Zhang et al.~\cite{zhang2020du2net} leveraged dual DP cameras for supervised depth learning. 
Pan et al.~\cite{pan2021dual} developed a DP simulator and jointly addressed depth estimation and deblurring. 
Xin et al.~\cite{xin2021defocus} formulated depth inference from defocus maps under unsupervised optimization. 
Kim et al.~\cite{kim2023spatio} enforced bidirectional disparity consistency across spatial and focal domains, 
achieving improved performance on synthetic and real datasets. 
Yu et al.~\cite{yu2025all} proposed all-directional disparity estimation for real-world QPD images, 
emphasizing the influence of noise, blur, and directional effects. 
Ghanekar et al.~\cite{ghanekar2024passive} further explored joint optical-computational design via coded apertures to enhance disparity observability.
Lee et al.~\cite{doehyung2025fmdp} explored incorporating foundation-model features to improve the robustness of DP disparity estimation.

Despite these advances, the extremely small baseline of DP imaging fundamentally limits disparity observability. 
In textureless, planar, or low-light scenarios, disparity signals become unstable, 
leading to degradation in weak-disparity regions. 
This intrinsic constraint suggests that relying primarily on DP cues---even with improved modeling or optical design---remains insufficient for stable and reliable metric depth estimation in complex real-world scenes.

\subsection{Monocular Depth Estimation Foundation Models}

Monocular depth estimation (MDE) seeks to recover dense geometry from a single RGB image and is inherently ill-posed. 
Early learning-based methods relied primarily on convolutional neural networks (CNNs). 
Eigen et al.~\cite{eigen2014depth} introduced a multi-scale architecture that predicts global structure followed by local refinement. 
Subsequent works improved stability through loss design and distribution modeling, 
including LRC~\cite{godard2017unsupervised} and DORN~\cite{fu2018deep}. 
Multi-task frameworks such as PAD-Net~\cite{xu2018pad} incorporated auxiliary geometric cues, 
while MiDaS~\cite{ranftl2020towards} and LeReS~\cite{yin2021learning} improved cross-dataset generalization via scale-invariant training and refinement strategies. 
Although effective under controlled settings, CNN-based models are limited in capturing long-range dependencies due to their local receptive fields.

The adoption of ViT has shifted MDE toward the foundation model paradigm. 
ViT~\cite{dosovitskiy2020VIT} models global dependencies through self-attention, 
and DPT~\cite{ranftl2021vision} first integrated ViT into dense depth prediction, 
marking a transition from CNN-based to Transformer-based architectures. 
Building on this paradigm, subsequent works have scaled model capacity and data diversity to enhance generalization.

Depth Anything~\cite{yang2024depth} employed a vision foundation encoder with teacher-student distillation, 
leveraging unlabeled real-world images for robust zero-shot depth prediction. 
DAV2~\cite{yang2024depthV2} further improved structural fidelity by training a large teacher on high-quality synthetic depth data before distillation to real images. 
MoGe~\cite{wang2025moge} and UniDepth~\cite{piccinelli2024unidepth} improved geometric consistency and metric depth estimation via supervision design and scale modeling, respectively. 
These ViT-based models demonstrate strong structural reasoning and cross-domain generalization through large-scale training and global attention mechanisms.

Diffusion models have also been explored for MDE. 
Marigold~\cite{ke2024repurposing} reformulated denoising generation as conditional depth prediction by fine-tuning Stable Diffusion. 
However, diffusion-based approaches require iterative inference and incur higher computational cost. 
In contrast, discriminative ViT-based foundation models achieve competitive structural modeling with significantly higher efficiency, 
making them more practical for real-world deployment.

%% file: main_sec/3_method.tex
\section{Method}
\label{sec:method}

\begin{figure*}[ht]
    \centering
    \includegraphics[width=1\linewidth]{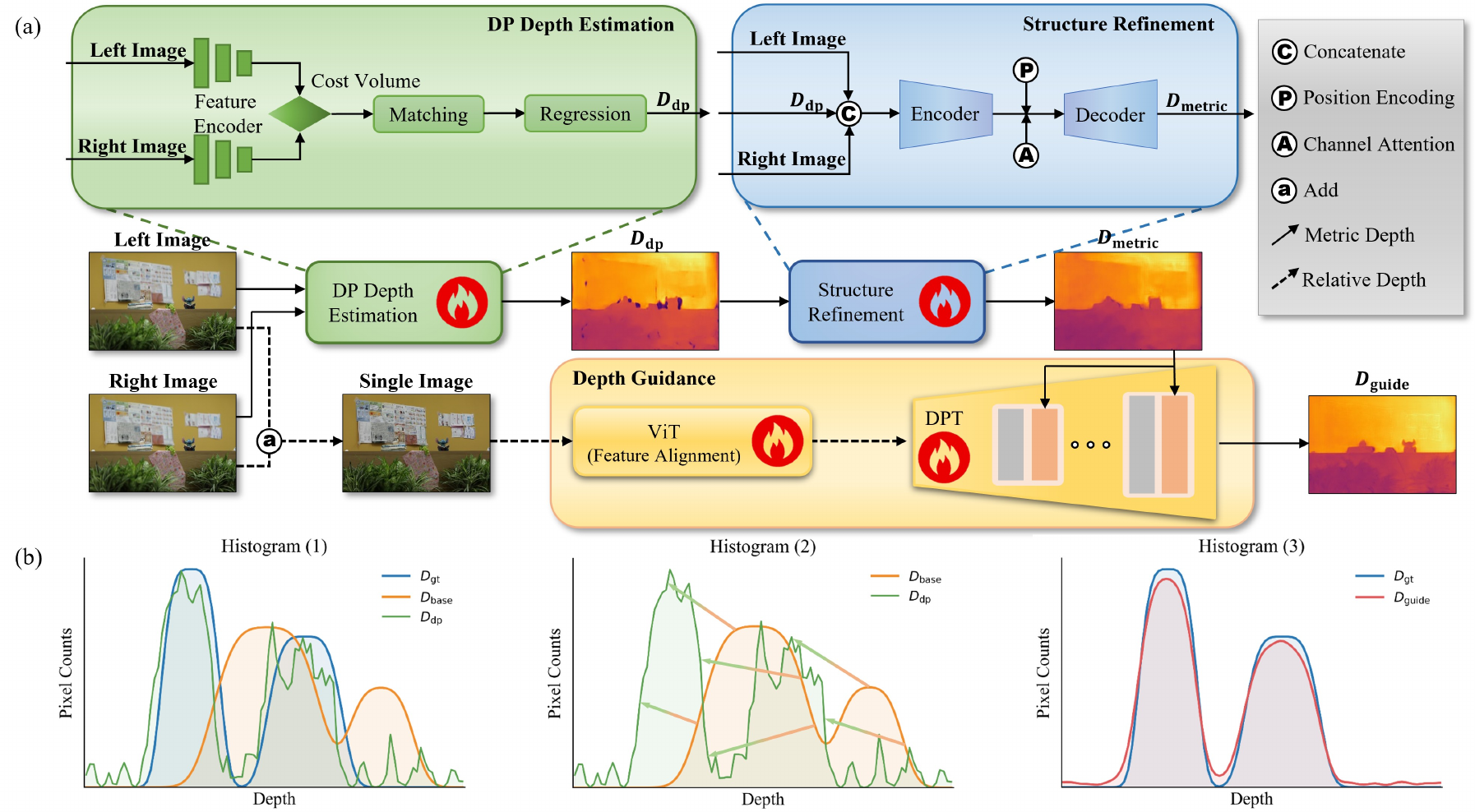}
    \vspace{-4mm}
    \caption{\label{fig:m_1}
(a) Overall framework of FoundDP. 
DP sub-aperture images are used to estimate metric depth $D_{\text{dp}}$, which is refined to $D_{\text{metric}}$ for improved geometric consistency. 
In parallel, a monocular depth foundation model extracts structural priors. 
A depth guidance module then uses metric depth to guide the structural features and produce the final depth $D_{\text{guide}}$.
(b) Illustration of depth histogram analysis.
(1) Compared to ground truth $D_{\text{gt}}$, foundation depth $D_{\text{base}}$ exhibits peak shifts, while DP depth $D_{\text{dp}}$ contains significant noise.
(2) Metric depth guides foundation priors to correct structural errors over noisy DP depth.
(3) The final depth $D_{\text{guide}}$ closely aligns with the ground-truth distribution, demonstrating improved structural and metric accuracy.
    }
    \vspace{-1mm}
\end{figure*}

\subsection{Overall Framework}

DP imaging provides metric depth cues grounded in explicit physical disparity induced by sub-aperture separation. 
However, the extremely small effective baseline limits disparity observability, particularly in textureless, low-contrast, or downsampled regions, leading to structural discontinuities and depth collapse. 
To address this limitation, we reconcile \emph{metric observability} from DP imaging with \emph{foundation-scale structural reasoning} from monocular depth models. 
DP depth serves as a metric anchor, while global structural priors are introduced to restore geometric coherence in weak-disparity regions.

As illustrated in Fig.~\ref{fig:m_1}(a), FoundDP consists of three stages: 
(1) DP Depth Estimation Module (DDE) producing initial metric depth $D_{\text{dp}}$; 
(2) Structure Refinement Module (SR) generating $D_{\text{metric}}$; 
(3) Depth Guidance Module (DG) using metric DP depth to guide ViT-based structural features, producing the final prediction $D_{\text{guide}}$.

To further analyze the complementary properties of DP depth and foundation-based depth priors, Fig.~\ref{fig:m_1}(b) presents depth histogram comparisons.
As shown in Fig.~\ref{fig:m_1}(b)(1), foundation depth $D_{\text{base}}$ captures coherent global structure but exhibits shifted peak locations relative to ground truth, indicating metric scale inconsistency, whereas DP depth $D_{\text{dp}}$ preserves approximate scale but contains substantial noise and structural instability.
Fig.~\ref{fig:m_1}(b)(2) further illustrates that metric DP depth guides foundation priors to suppress noise and improve distribution consistency over DP depth.
As a result, the final guided depth $D_{\text{guide}}$ closely matches the ground-truth depth distribution, as shown in Fig.~\ref{fig:m_1}(b)(3), demonstrating that our framework effectively combines metric observability with global structural consistency.

The overall formulation is:
\begin{equation}
\left\{
\begin{alignedat}{2}
D_{\text{dp}}     &= \mathcal{F}_{\text{DDE}}(I_L, I_R), \\
D_{\text{metric}} &= \mathcal{F}_{\text{SR}}(I_L, I_R, D_{\text{dp}}), \\
D_{\text{guide}} &= \mathcal{F}_{\text{DG}}\!\left(
D_{\text{metric}}, 
\Phi_{\text{ViT}}\!\left(\tfrac{I_L + I_R}{2}\right)
\right),
\end{alignedat}
\right.
\end{equation}
where $\Phi_{\text{ViT}}$ denotes structural features extracted by the foundation encoder, and $\mathcal{F}_\mathrm{DDE}$, $\mathcal{F}_\mathrm{SR}$, and $\mathcal{F}_\mathrm{DG}$ represent the DDE, SR, and DG respectively.

\subsection{DP Depth Estimation Module}

We first construct a DP-based network to generate an initial metric depth map $D_{\text{dp}}$ with explicit physical scale. 
Given left and right DP images $I_L, I_R \in \mathbb{R}^{3\times H\times W}$, a shared-weight encoder extracts multi-scale representations. 
The encoder integrates strided convolutions for receptive field enlargement with dilated convolutions and pooling branches to enhance contextual aggregation in textureless regions. 
Multi-scale features are fused at a unified resolution and compressed via $1\times1$ convolutions, producing matching features $F_L, F_R \in \mathbb{R}^{C\times H'\times W'}$, where $H' = H/4$ and $W' = W/4$.

A 3D cost volume~\cite{pan2021dual,he2025simulating} is constructed through explicit displacement concatenation. 
Instead of conventional single-sided search, we define a symmetric displacement hypothesis set centered at zero disparity, $d \in [-D/2, D/2]$. 
For each hypothesis, aligned left and right features are concatenated along the channel dimension, forming a volume of size $2C \times D \times H' \times W'$. 
This symmetric design aligns with the optical symmetry of DP sub-apertures and reduces systematic bias during depth learning. 
The volume is regularized using a 3D convolutional network to jointly model spatial and disparity consistency.

To obtain continuous predictions, we adopt Softmin-based regression. 
The regularized volume is upsampled to the target resolution, and Softmin is applied along the disparity dimension to obtain a probability distribution. 
The final depth is computed as
\begin{equation}
D_{\text{dp}}(x,y)
=
\mathbb{E}_{d \sim P(d \mid x,y)}[d],
\end{equation}
where
\begin{equation}
P(d \mid x,y)
=
\frac{\exp(-C(x,y,d))}
{\sum_{d'} \exp(-C(x,y,d'))},
\end{equation}
and $\mathbb{E}$ denotes expectation over valid pixels.
This strategy avoids discrete quantization artifacts and produces smoother predictions in textureless and low-contrast regions. 
Nevertheless, because this module relies primarily on local disparity cues, structural degradation may persist in weak-disparity areas, motivating subsequent refinement and guidance.

\subsection{Structure Refinement Module}

Although the DP module provides physically grounded depth $D_{\text{dp}}$, its reliance on local disparity cues makes it vulnerable to structural degradation in regions with weak disparity observability, such as textureless or planar areas.
To address this limitation while preserving metric scale, we introduce a structure refinement network that predicts a residual structural correction conditioned on both image appearance and the initial DP depth.

Specifically, the refinement module estimates a correction field:
\begin{equation}
\Delta D
=
\mathcal{SR}
\left(
I_L, I_R, D_{\text{dp}}
\right),
\qquad
D_{\text{metric}}
=
D_{\text{dp}}
+
\Delta D,
\end{equation}
where $\mathcal{SR}(\cdot)$ denotes the refinement network and $\Delta D$ represents the learned structural correction.
This residual formulation preserves the physically grounded metric scale while compensating for structural errors caused by insufficient disparity observability.

To implement $\mathcal{SR}(\cdot)$, the network jointly processes $I_L$, $I_R$, and $D_{\text{dp}}$ using a hierarchical encoder-decoder architecture~\cite{pan2024weakly}.
The inputs are concatenated along the channel dimension and passed through multi-scale residual convolutional blocks, enabling progressive aggregation of local image evidence and global structural context.
A channel attention~\cite{wang2020eca,bastidas2019channel} and 2D positional encoding~\cite{wu2021rethinking} mechanism is applied at high-resolution stages to emphasize geometrically informative responses and suppress unreliable regions.
The decoder reconstructs the refined depth in a residual manner, ensuring structural continuity while maintaining metric consistency.

The resulting depth $D_{\text{metric}}$ provides structurally refined metric depth and serves as the geometric foundation for the subsequent guidance.

\subsection{Depth Guidance Module}

\begin{wrapfigure}{l}{0.5\textwidth}
    \centering
    \includegraphics[width=0.5\columnwidth]{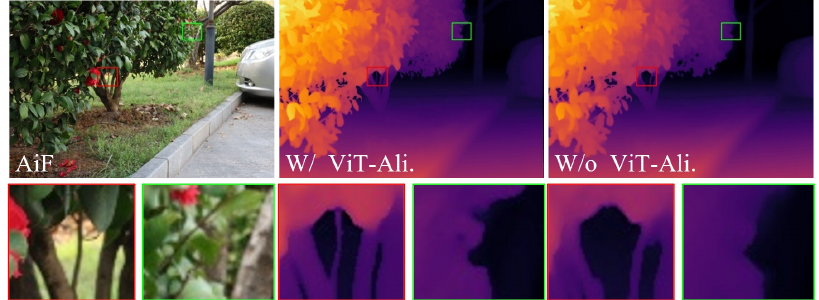}
    \caption{\label{fig:m_2}
    Effect of ViT alignment under defocus degradation.
    Aligned ViT features yield more structurally consistent predictions in blurred regions.
    }
\end{wrapfigure}

\noindent\textbf{ViT Alignment under DP Degradation.}  
After obtaining refined metric depth, we introduce monocular structural priors from a pretrained ViT~\cite{dosovitskiy2020VIT}. 
Unlike prior fusion approaches that assume clean representations, DP defocus and optical blur attenuate high-frequency content and induce structural bias in ViT features, as illustrated in \cref{fig:m_2}. 
Directly fusing such degraded features can propagate structural inconsistencies and compromise depth reliability.

To mitigate this degradation, we apply explicit ViT feature alignment prior to restoring structural consistency.
Specifically, paired clear RGB images and their DP-degraded counterparts are processed by a shared ViT encoder, and a feature-space alignment constraint is imposed to reduce representation discrepancies caused by defocus blur.
This alignment encourages degraded features to recover structurally consistent representations, providing more reliable global priors.

Formally, we supervise the feature alignment with the following objective:
\begin{equation}
\mathcal{L}_{\text{align}}
=
\sum_l
\left\|
\Phi_l(I_{\text{clear}})
-
\Phi_l(I_{\text{blur}})
\right\|_2^2,
\end{equation}
where $\Phi_l$ denotes the ViT feature at level $l$.
This objective explicitly enforces representation consistency between defocused and clean inputs, mitigating defocus-induced structural bias.

\noindent\textbf{Depth Guidance with Metric Conditioning.}  
Inspired by depth-prompt strategies~\cite{lin2025prompting,ye2025prompthaze,wang2025depth,jiang2025defom,wen2025foundationstereo}, 
we treat the refined DP metric depth as an explicit geometric condition during depth guidance, preserving metric scale while leveraging foundation-level structural priors.

The RGB image is first encoded by the ViT into multi-level token features, which are reshaped into multi-scale feature maps $\{F_1, F_2, F_3, F_4\}$~\cite{yang2024depthV2}. 
These features are progressively decoded in a DPT-style architecture. 
At each decoding stage, let $X$ denote the intermediate feature map at the current scale. 
The aligned and normalized metric depth $\hat{D}$ is injected as a spatial condition to guide features.

Specifically, feature updates are formulated as:
\begin{equation}
    X' = X + \phi(X) + g(\hat{D}) \odot \psi(\hat{D}),
\end{equation}
where $\phi(\cdot)$ denotes a residual feature transformation applied to $X$, 
$\psi(\cdot)$ extracts metric-conditioned modulation signals from $\hat{D}$, 
and $g(\cdot)$ is an adaptive gating function that controls the spatial influence of the metric condition.
The element-wise modulation $\odot$ enables selective enhancement of geometrically reliable regions.
This design encourages the network to rely on DP-derived metric cues in high-confidence areas while allowing foundation-based structural priors to dominate in weak-disparity regions, thereby reducing error propagation.

\subsection{Loss Function}

We adopt a unified Smooth L1 loss in logarithmic depth space to supervise all three stages:
\begin{equation}
    \mathcal{L} = \frac{1}{N} \sum_i \text{SmoothL1}\big(\log_{10} D^{(i)}, \log_{10} D_{\text{gt}}^{(i)}\big),
\end{equation}
where $D^{(i)}$ denotes predictions from DP estimation, refinement, and guidance, and $D_{\text{gt}}$ is the ground truth.
We do not adopt scale-invariant log losses (e.g., SiLog~\cite{eigen2014depth}), 
as dual-pixel has the capability to recover metric depth, where preserving absolute scale is necessary for physical consistency.

%% file: main_sec/4_exp.tex
\section{Experiments}
\label{sec:exp}

\subsection{Implementation Details}

Our framework uses DAV2 as the foundation model and is implemented in PyTorch, trained on a single NVIDIA RTX 4090D GPU.
All trainable modules are optimized using Adam with an initial learning rate of $1\times10^{-4}$ and trained for 100 epochs per stage.
Following common practice, we resize all images to a fixed resolution to ensure computational efficiency and consistent evaluation across datasets.
In our experiments, images are resized to $512\times768$ for both training and evaluation.

\noindent\textbf{Training Strategy.}
Given the cascaded design and heterogeneous objectives of different modules, we adopt a stage-wise optimization scheme for stable convergence.
First, the Dual-Pixel Depth Estimation Module is trained independently to establish reliable metric-scale predictions.
The Structure Refinement Module is then trained with DDE frozen to enhance geometric continuity in weak-disparity regions.
Next, the pretrained ViT encoder of Depth Guidance Module is introduced and refined under DP imaging conditions to mitigate representation degradation caused by defocus blur.
Finally, DDE, SR, and the ViT encoder are fixed, and only the DPT is optimized.
This staged training preserves metric consistency while enabling stable integration of global structural priors.






\begin{figure*}[ht]
    \centering
    \includegraphics[width=1\linewidth]{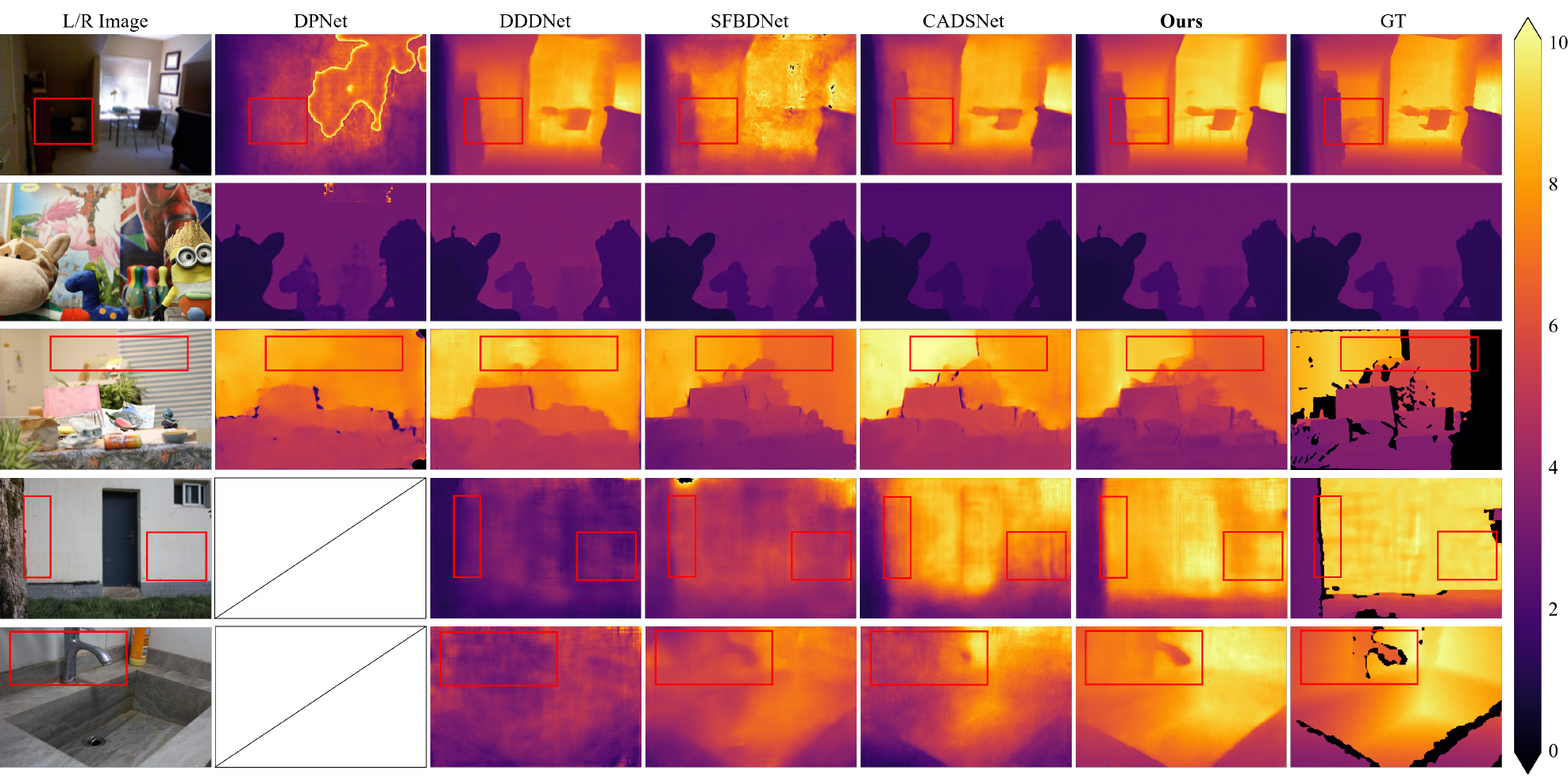}
    \vspace{-3mm}
    \caption{\label{fig:e_1}
    Qualitative comparison.
    \textbf{Rows 1-3:} Results on representative DP datasets.
    Our method preserves sharper structural boundaries and more coherent geometry,
    especially in weak-disparity regions (highlighted in red).
    \textbf{Rows 4-5:} Results on the downsampling dataset.
    Our model remains robust under disparity attenuation,
    whereas competing methods exhibit structural degradation.
    }
    \vspace{-1mm}
\end{figure*}

\subsection{Datasets and Evaluation Metrics}

\noindent\textbf{Datasets.}
We evaluate on both synthetic and real-world DP datasets. 
For each dataset, the training and test sets are generated or captured under the same imaging model or camera setup. 

\textbf{Synthetic:} NYUData~\cite{silberman2012indoor}.  
Since the synthetic dataset contains only single-view RGB images and depth maps, we generate DP image pairs using a 
ray-tracing-based DP simulator~\cite{he2025simulating}, which produces pixel-wise left/right point spread functions and corresponding DP observations.

\textbf{Real-world:} DP2020~\cite{punnappurath2020modeling}, DP5K~\cite{li2023learning}, and DP2019~\cite{garg2019learning}.
The real datasets provide DP left-right image pairs with corresponding depth annotations.

To study robustness against disparity attenuation, we additionally construct a downsampling dataset named DPDown70 using a Canon EOS R6 Mark II camera equipped with an RF 50mm lens at an aperture of f/8,
including $\sim$1k training images and $\sim$70 test samples.
Original $4000\times6000$ images are resized to $512\times768$.

To ensure consistent metric evaluation, all valid depths are linearly mapped to a unified physical range of 1-10 meters.
Values outside this range are ignored.
For datasets providing disparity only, depth is computed as:

\begin{equation}
D = \frac{1}{a + b d}, \quad
a = \frac{1}{D_{\max}}, \quad
b = \frac{1}{D_{\min}} - \frac{1}{D_{\max}},
\end{equation}
with $D_{\min}=1.0$ m and $D_{\max}=10.0$ m.

\noindent\textbf{Metrics.}
We evaluate depth quality from three perspectives:

\textbf{Affine-Invariant Error (AI)}~\cite{garg2019learning,pan2021dual,kim2023spatio,yu2025all}: AI(1) and AI(2) measure mean absolute and squared errors after affine alignment.  

\textbf{Rank Consistency}~\cite{kim2023spatio,yu2025all}: Spearman correlation $\rho_s$.  

\textbf{Threshold Accuracy}~\cite{pan2021dual,he2025simulating}: $\delta < 1.25$ (Acc-1) and $\delta < 1.25^2$ (Acc-2).

These complementary metrics jointly assess structural fidelity, relative ordering, and metric accuracy.

\begin{table*}[h]
    \centering
    \caption{Quantitative test on NYUData~\cite{silberman2012indoor}, DP2020~\cite{punnappurath2020modeling}, DP5K~\cite{li2023learning} and DP2019~\cite{garg2019learning}.}
\label{tab:e1}
\vspace{-1mm} 
\resizebox{\columnwidth}{!}{
\begin{tabular}{l|ccccc|ccccc}
    \toprule
\multirow{2}{*}{Method} & \multicolumn{5}{c|}{NYUData~\cite{silberman2012indoor}}                                                                          & \multicolumn{5}{c}{DP2020~\cite{punnappurath2020modeling}}                                                                            \\
                        & AI(1)$\downarrow$ & AI(2)$\downarrow$ & $1-|\rho_s|$$\downarrow$ & Acc-1$\uparrow$ & Acc-2$\uparrow$ & AI(1)$\downarrow$ & AI(2)$\downarrow$ & $1-|\rho_s|$$\downarrow$ & Acc-1$\uparrow$ & Acc-2$\uparrow$ \\ \midrule
DPNet~\cite{garg2019learning}        & 0.9271            & 1.2896            & 0.2179                    & 0.2925          & 0.5777          & 0.3509            & 0.5629            & 0.1649                    & 0.7521          & 0.8180          \\
SFBDNet~\cite{kim2023spatio}         & 0.5330            & 0.8576            & 0.1489                    & 0.6579          & 0.9174          & 0.0633            & 0.2095            & {\ul 0.0241}              & {\ul 0.9963}    & 0.9987          \\
DDDNet~\cite{pan2021dual}            & 0.2271            & 0.3333            & 0.0299                    & {\ul 0.9520}    & {\ul 0.9953}    & 0.0851            & 0.2564            & 0.1138                    & 0.6934          & 0.9757          \\
CADSNet~\cite{ghanekar2024passive}   & {\ul 0.2018}      & {\ul 0.2931}           & {\ul 0.0248}              & 0.9271          & 0.9933          & {\ul 0.0162}      & {\ul 0.0424}      & 0.0467                    & 0.8153          & {\ul 0.9997}    \\
\textbf{Ours}                        & \textbf{0.1475}   & \textbf{0.2368}   & \textbf{0.0178}           & \textbf{0.9661} & \textbf{0.9988} & \textbf{0.0110}   & \textbf{0.0280}   & \textbf{0.0232}           & \textbf{0.9998} & \textbf{1.0000} \\ 
\midrule
\multirow{2}{*}{Method} & \multicolumn{5}{c|}{DP5K~\cite{li2023learning}}                                                                             & \multicolumn{5}{c}{DP2019~\cite{garg2019learning}}                                                                            \\
                        & AI(1)$\downarrow$ & AI(2)$\downarrow$ & $1-|\rho_s|$$\downarrow$ & Acc-1$\uparrow$ & Acc-2$\uparrow$ & AI(1)$\downarrow$ & AI(2)$\downarrow$ & $1-|\rho_s|$$\downarrow$ & Acc-1$\uparrow$ & Acc-2$\uparrow$ \\ \midrule
DPNet~\cite{garg2019learning}      & 0.4564            & 0.9296            & 0.3318                    & 0.6675          & 0.8903          & 0.1613            & 0.3211            & 0.5884                    &   0.4311          & 0.5807  \\
SFBDNet~\cite{kim2023spatio}       & 0.3406            & 0.8228            & 0.1131                    & 0.7757          & 0.8973          & 0.1549            & 0.3087            & 0.5223                    & 0.6312          & 0.7974          \\
DDDNet~\cite{pan2021dual}          & 0.3617            & 0.8279            & 0.1706                    & {\ul 0.7849}    & {\ul 0.9088}    & 0.1473            & 0.2935            & 0.4539                    & 0.6820          & 0.9031          \\
CADSNet~\cite{ghanekar2024passive} & {\ul 0.2680}      & {\ul 0.7854}           & \textbf{0.0977}           & 0.6716          & 0.8773          & {\ul 0.1313}      & {\ul 0.2807}      & {\ul 0.3491}              & {\ul 0.7335}    & {\ul 0.9277}          \\
\textbf{Ours}                      & \textbf{0.2579}   & \textbf{0.7562}   & {\ul 0.0989}              & \textbf{0.7976} & \textbf{0.9251} & \textbf{0.1168}   & \textbf{0.2622}   & \textbf{0.3224}           & \textbf{0.8571} & \textbf{0.9404} \\ 
\bottomrule
\end{tabular}
}
\end{table*}

\begin{table*}[h]
    \centering
    \caption{
    Quantitative evaluation \emph{in weak regions} on NYUData~\cite{silberman2012indoor}, DP2020~\cite{punnappurath2020modeling}, DP5K~\cite{li2023learning} and DP2019~\cite{garg2019learning}.
}
\label{tab:e2}
\vspace{-1mm} 
\resizebox{\columnwidth}{!}{
\begin{tabular}{l|ccccc|ccccc}
    \toprule
\multirow{2}{*}{Method} & \multicolumn{5}{c|}{NYUData~\cite{silberman2012indoor}}                                                                          & \multicolumn{5}{c}{DP2020~\cite{punnappurath2020modeling}}                                                                            \\
                        & AI(1)$\downarrow$ & AI(2)$\downarrow$ & $1-|\rho_s|$$\downarrow$ & Acc-1$\uparrow$ & Acc-2$\uparrow$ & AI(1)$\downarrow$ & AI(2)$\downarrow$ & $1-|\rho_s|$$\downarrow$ & Acc-1$\uparrow$ & Acc-2$\uparrow$ \\ 
                        \midrule
DPNet~\cite{garg2019learning}                   & 0.8512            & 1.1677            & 0.2340                    & 0.2544          & 0.5225          & 0.3389            & 0.5938            & 0.1931                    & 0.8087          & 0.8706          \\
SFBDNet~\cite{kim2023spatio}                 & 0.5038            & 0.7853            & 0.1698                    & 0.6579          & 0.9243          & 0.0622            & 0.2155            & {\ul 0.0258}              & {\ul 0.9942}    & 0.9979          \\
DDDNet~\cite{pan2021dual}                  & 0.2504            & 0.3471            & 0.0438                    & {\ul 0.9391}    & 0.9940          & 0.0717            & 0.2938            & 0.1397                    & 0.7220          & 0.9933          \\
CADSNet~\cite{ghanekar2024passive}                 & {\ul 0.1949}      & {\ul 0.2725}      & {\ul 0.0320}              & 0.9276          & {\ul 0.9947}    & {\ul 0.0113}      & {\ul 0.0416}      & 0.0437                    & 0.7472          & {\ul 0.9995}    \\
\textbf{Ours}                    & \textbf{0.1280}   & \textbf{0.1872}   & \textbf{0.0180}           & \textbf{0.9675} & \textbf{0.9995} & \textbf{0.0044}   & \textbf{0.0157}   & \textbf{0.0113}           & \textbf{1.0000} & \textbf{1.0000} \\ 
\midrule
\multirow{2}{*}{Method} & \multicolumn{5}{c|}{DP5K~\cite{li2023learning}}                                                                             & \multicolumn{5}{c}{DP2019~\cite{garg2019learning}}                                                                            \\
                        & AI(1)$\downarrow$ & AI(2)$\downarrow$ & $1-|\rho_s|$$\downarrow$ & Acc-1$\uparrow$ & Acc-2$\uparrow$ & AI(1)$\downarrow$ & AI(2)$\downarrow$ & $1-|\rho_s|$$\downarrow$ & Acc-1$\uparrow$ & Acc-2$\uparrow$ \\ 
                        \midrule
DPNet~\cite{garg2019learning}                   & 0.3877            & 0.8228            & 0.3746                    & 0.6879          & 0.8962          & 0.1955            & 0.3685            & 0.6234                    &  0.4107          & 0.6061         \\
SFBDNet~\cite{kim2023spatio}                 & 0.3147            & 0.7439            & 0.2986                    & {\ul 0.7592}    & {\ul 0.8996}    & 0.1874            & 0.3542            & 0.5490                    & 0.5740          & {\ul 0.7291}    \\
DDDNet~\cite{pan2021dual}                  & 0.3393            & 0.7736            & 0.3079                    & 0.7551          & 0.8936          & 0.1894            & 0.3553            & 0.6033                    & 0.5176          & 0.6693          \\
CADSNet~\cite{ghanekar2024passive}                 & {\ul 0.2781}      & {\ul 0.7283}      & {\ul 0.2526}              & 0.6965          & 0.8738          & {\ul 0.1834}      & {\ul 0.3489}      & {\ul 0.5226}              &    {\ul 0.6143}    & 0.7098       \\
\textbf{Ours}                    & \textbf{0.2441}   & \textbf{0.7019}   & \textbf{0.2286}           & \textbf{0.7777} & \textbf{0.9264} & \textbf{0.1795}   & \textbf{0.3432}   & \textbf{0.5067}           & \textbf{0.6229} & \textbf{0.7552} \\ 
\bottomrule
\end{tabular}
}
\end{table*}

\subsection{Comparison with DP Methods}

We compare against representative DP-based methods:
DPNet~\cite{garg2019learning},
DDDNet~\cite{pan2021dual},
SFBDNet~\cite{kim2023spatio},
and CADSNet~\cite{ghanekar2024passive}.
All methods are evaluated under identical preprocessing and metric protocols.
The best results are highlighted in bold and the second-best results are underlined in the table.

\noindent\textbf{General Comparison.}
As shown in Table~\ref{tab:e1},
our method achieves the best or second-best results.
Notably, we consistently obtain the lowest affine-invariant errors and the highest threshold accuracies.
Compared to earlier DPNet and DDDNet, our approach substantially reduces overall error.
Against recent methods such as SFBDNet and CADSNet,
our framework further improves metric accuracy while preserving structural consistency,
demonstrating that global structural priors effectively compensate for DP observability limitations.
Visual comparisons in Fig.~\ref{fig:e_1} (Rows 1-3) confirm that our method produces smoother planar surfaces,
more stable boundaries, and fewer depth holes.

\noindent\textbf{Performance in Weak-Disparity Regions.}
Weak-disparity regions are identified using gradient-based analysis.
Sobel responses are smoothed, thresholded, and refined with morphological operations to obtain stable masks. More details are listed in supplementary materials.

\begin{wraptable}{r}{0.5\textwidth}
    \centering
    \caption{
    Quantitative evaluation on our \emph{Downsampling Dataset} DPDown70.
}
\label{tab:e3}
\resizebox{0.5\columnwidth}{!}{
\begin{tabular}{l|ccccc}
\toprule
\multirow{2}{*}{Method} & \multicolumn{5}{c}{DPDown70}                                                                                                                                                                          \\
                        & \multicolumn{1}{c}{AI(1)$\downarrow$} & \multicolumn{1}{c}{AI(2)$\downarrow$} & \multicolumn{1}{c}{$1-|\rho_s|$$\downarrow$} & \multicolumn{1}{c}{Acc-1$\uparrow$} & \multicolumn{1}{c}{Acc-2$\uparrow$} \\ 
\midrule
SFBDNet~\cite{kim2023spatio}                 & 0.6458                                & 1.0073                                & 0.2478                                        & 0.4113                              & 0.7309                              \\
DDDNet~\cite{pan2021dual}                 & 0.6379                                & 0.9120                                & 0.2215                                        & 0.3524                              & 0.6746                              \\
CADSNet~\cite{ghanekar2024passive}                 & {\ul 0.4003}                          & {\ul 0.6281}                          & {\ul 0.1238}                                  & {\ul 0.5047}                        & {\ul 0.8074}                        \\
\textbf{Ours}                    & \textbf{0.2482}                       & \textbf{0.4078}                       & \textbf{0.0443}                               & \textbf{0.7035}                     & \textbf{0.9163}                     \\ 
\bottomrule
\end{tabular}
}
\end{wraptable}

Table~\ref{tab:e2} shows that competing DP methods degrade significantly in weak regions,
while our method maintains substantially lower AI errors and higher threshold accuracies.
The improvement gap is more pronounced than in global evaluation,
indicating that foundation-based structural priors effectively compensate when local DP correspondence fails.

\noindent\textbf{Downsampling Robustness.}
Under resolution reduction (Table~\ref{tab:e3}),
most DP methods suffer severe performance degradation.
In contrast, our method maintains strong metric accuracy and structural coherence.
The qualitative results in Fig.~\ref{fig:e_1} (Rows 4-5) further demonstrate that our approach preserves geometric contours despite disparity attenuation.



\noindent\textbf{Disparity Observability Analysis.}
The weak-disparity region analysis above evaluates performance under naturally occurring observability variations.
To further investigate the impact of disparity observability in a controlled manner, we progressively reduce disparity observability via downsampling.

In DP imaging, disparity cues arise from subpixel correspondence between dual sub-aperture views.
Downsampling suppresses fine-scale intensity variations and reduces effective gradient strength, thereby weakening correspondence signals required for depth estimation.
As a result, disparity observability is inversely related to the downsampling factor, allowing resolution reduction to serve as a continuous observability proxy.

\begin{figure}[t]
\centering

\begin{minipage}[t]{0.48\textwidth}
\centering
\includegraphics[width=\linewidth]{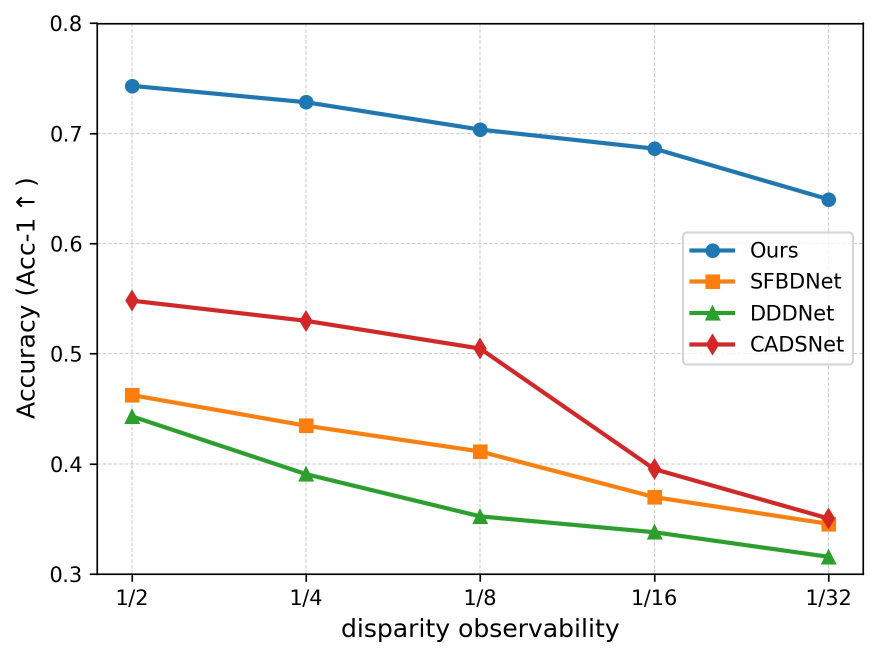}
\caption{\label{fig:e_2}
Depth accuracy versus disparity observability.
Our method shows improved robustness under reduced observability.
}
\end{minipage}
\hfill
\begin{minipage}[t]{0.48\textwidth}
\centering
\includegraphics[width=\linewidth]{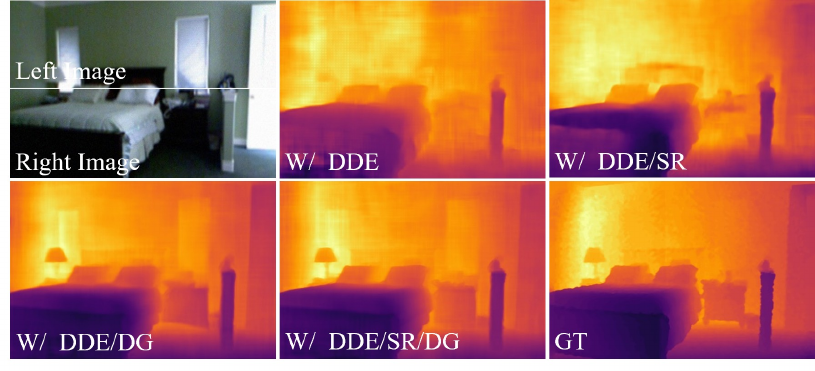}
\caption{\label{fig:e_3}
Ablation results.
Progressive integration of SR and DG improves boundary sharpness,
structural continuity,
and weak-region stability.
}
\end{minipage}

\vspace{-2mm}
\end{figure}

We evaluate depth accuracy under the Acc-1 criterion as a function of disparity observability.
As shown in Fig.~\ref{fig:e_2}, all methods exhibit performance degradation as observability decreases, confirming the fundamental dependence of DP depth estimation on reliable disparity cues.
However, existing DP methods degrade rapidly under reduced observability, indicating their strong reliance on local correspondence signals.
In contrast, our method maintains consistently higher accuracy and exhibits a more gradual degradation trend, demonstrating improved robustness when disparity observability becomes limited.

\subsection{Ablation Studies}
\noindent\textbf{Module Contribution Analysis.} 
Table~\ref{tab:e4} reports progressive integration of core modules.
Using DDE alone provides stable metric depth but leaves structural artifacts.
Adding SR reduces local discontinuities and improves affine-invariant accuracy,
confirming its role in repairing depth holes.
Incorporating DG further yields the largest performance gain,
demonstrating that global structural priors effectively stabilize geometry when disparity observability deteriorates.
Qualitative results in Fig.~\ref{fig:e_3} match the quantitative trends:
depth discontinuities are progressively suppressed,
and structural coherence improves as modules are added.

\begin{table}[t]
\centering

\begin{minipage}{0.5\textwidth}
\centering
\caption{Ablation study on three modules.}
\vspace{-2mm} 
\label{tab:e4}
\resizebox{\linewidth}{!}{
\begin{tabular}{ccc|ccccc}
\toprule
\multicolumn{3}{c|}{Components} & \multicolumn{5}{c}{NYUData} \\
DDE & SR & DG & AI(1)$\downarrow$ & AI(2)$\downarrow$ & $1-|\rho_s|$$\downarrow$ & Acc-1$\uparrow$ & Acc-2$\uparrow$ \\
\midrule
$\checkmark$ & & & 0.2463 & 0.3764 & 0.0332 & 0.9297 & 0.9610 \\
$\checkmark$ & $\checkmark$ & & 0.2312 & 0.3486 & 0.0298 & 0.9491 & 0.9715 \\
$\checkmark$ & & $\checkmark$ & 0.1865 & 0.2871 & 0.0220 & 0.9348 & 0.9883 \\
$\checkmark$ & $\checkmark$ & $\checkmark$ & \textbf{0.1475} & \textbf{0.2368} & \textbf{0.0178} & \textbf{0.9661} & \textbf{0.9988} \\
\bottomrule
\end{tabular}
}
\end{minipage}
\hfill
\begin{minipage}{0.49\textwidth}
\centering
\caption{Ablation study on ViT feature alignment.}
\label{tab:e5}
\vspace{-2mm}
\resizebox{\linewidth}{!}{
\begin{tabular}{cc|ccccc}
\toprule
\multicolumn{2}{c|}{Components} & \multicolumn{5}{c}{DP5K} \\
Base & ViT-Ali. & AI(1)$\downarrow$ & AI(2)$\downarrow$ & $1-|\rho_s|$$\downarrow$ & Acc-1$\uparrow$ & Acc-2$\uparrow$ \\
\midrule
$\checkmark$ & & 0.2741 & 0.7993 & 0.1042 & 0.7839 & 0.9197 \\
$\checkmark$ & $\checkmark$ & \textbf{0.2579} & \textbf{0.7562} & \textbf{0.0989} & \textbf{0.7976} & \textbf{0.9251} \\
\bottomrule
\end{tabular}
}
\end{minipage}
\vspace{-2mm} 
\end{table}

\noindent\textbf{Effect of ViT feature alignment.}
Table~\ref{tab:e5} evaluates the impact of the proposed ViT feature alignment strategy.
Directly incorporating foundation features improves performance, confirming the benefit of global structural priors.
However, performance remains constrained due to representation degradation caused by DP blur, which introduces bias in the extracted ViT features; introducing feature alignment mitigates this effect and improves performance.

\begin{wraptable}{l}{0.3\textwidth}
    \centering
    \caption{
    ViT feature alignment analysis.
}
\label{tab:e6}
\resizebox{\linewidth}{!}{
\begin{tabular}{l|c}
\toprule
ViT Feature             & \multicolumn{1}{l}{Cosine Similarity$\uparrow$} \\ \midrule
Origin & 0.8737                                          \\
Refined & 0.9454                                          \\ \bottomrule
\end{tabular}
}
\end{wraptable}

To further analyze this effect, Table~\ref{tab:e6} measures features before and after alignment using cosine similarity to clean-image features as reference.
DP blur degrades ViT feature representations, while the proposed alignment strategy restores feature alignment.
By correcting this degradation-induced bias, the alignment enables more effective metric-guided integration with DP metric cues, leading to consistent improvements across all metrics in Table~\ref{tab:e5}.
This confirms that mitigating ViT representation degradation is essential for stable depth estimation under DP imaging.

Overall, the ablation results confirm that
(1) SR mitigates local geometric degradation,
(2) DG injects global structural reasoning,
and
(3) ViT feature alignment stabilizes representations under DP-induced blur.
Together, these components address the core limitation of weak disparity observability in DP depth estimation.
Lastly, complexity and runtime analysis further demonstrate that the proposed framework achieves these improvements with practical inference efficiency (see supplementary for details).

%% file: main_sec/5_conc.tex
\section{Conclusion and Discussion}
\label{sec:conc}

DP depth estimation is fundamentally constrained by weak disparity observability, 
causing structural degradation and unreliable predictions in weak-disparity regions. 
We address this limitation by reconciling physically grounded metric cues from DP imaging with global structural priors from monocular depth foundation models. 
Our FoundDP preserves metric scale through DP disparity while leveraging foundation representations to restore coherence where disparity becomes unreliable. 
Extensive experiments on synthetic and real-world benchmarks demonstrate consistent gains in structural fidelity and metric accuracy, particularly under weak-disparity and downsampling conditions.

Despite these advances, the framework inherently depends on observable disparity signals; performance may degrade when DP measurements are heavily corrupted by noise or extreme optical degradation. 
While foundation-based priors improve robustness, their adaptation to DP-specific imaging characteristics remains imperfect. 
Moreover, integrating a ViT-based encoder introduces additional computational overhead, limiting real-time deployment on resource-constrained devices. 
Nevertheless, incorporating foundation-level structural reasoning is essential for stabilizing metric DP depth under extremely small-baseline and low-disparity conditions, establishing a principled bridge between physical imaging constraints and large-scale representation learning.